\documentclass{article}

\usepackage{microtype}
\usepackage{graphicx}
\usepackage{subfigure}
\usepackage{booktabs} % for professional tables

\usepackage{hyperref}

\usepackage[accepted]{lib/icml2025}

\usepackage{amsmath}
\usepackage{amssymb}
\usepackage{mathtools}
\usepackage{amsthm}

\usepackage[capitalize,noabbrev]{cleveref}

\theoremstyle{plain}
\newtheorem{theorem}{Theorem}[section]

\newtheorem{corollary}[theorem]{Corollary}
\theoremstyle{definition}
\newtheorem{definition}[theorem]{Definition}

\theoremstyle{remark}
\newtheorem{remark}[theorem]{Remark}

\usepackage[textsize=tiny]{todonotes}

\begin{document}

\twocolumn[
\icmltitle{Skewed Memorization in Large Language Models: Quantification and Decomposition}

\icmlsetsymbol{equal}{*}

\begin{icmlauthorlist}
\icmlauthor{Hao Li}{col,equal}
\icmlauthor{Di Huang}{wustl,equal}
\icmlauthor{Ziyu Wang}{uci,equal}
\icmlauthor{Amir M. Rahmani}{uci}
\end{icmlauthorlist}

\icmlaffiliation{col}{Columbia University, USA}
\icmlaffiliation{wustl}{Washington University in St. Louis, USA}
\icmlaffiliation{uci}{University of California, Irvine, USA}

\icmlcorrespondingauthor{Hao Li}{hl3776@columbia.edu}
\icmlcorrespondingauthor{Di Huang}{di.huang@wustl.edu}
\icmlcorrespondingauthor{Ziyu Wang}{ziyuw31@uci.edu}
\icmlcorrespondingauthor{Amir M. Rahmani}{amirr1@hs.uci.edu}

\icmlkeywords{Machine Learning, ICML}

\vskip 0.3in
]



\printAffiliationsAndNotice{\icmlEqualContribution} % otherwise use the standard text.

\begin{abstract} 
Memorization in Large Language Models (LLMs) poses privacy and security risks, as models may unintentionally reproduce sensitive or copyrighted data. Existing analyses focus on average-case scenarios, often neglecting the highly skewed distribution of memorization. This paper examines memorization in LLM supervised fine-tuning (SFT), exploring its relationships with training duration, dataset size, and inter-sample similarity. By analyzing memorization probabilities over sequence lengths, we link this skewness to the token generation process, offering insights for estimating memorization and comparing it to established metrics. Through theoretical analysis and empirical evaluation, we provide a comprehensive understanding of memorization behaviors and propose strategies to detect and mitigate risks, contributing to more privacy-preserving LLMs.

\end{abstract}

\section{Introduction}

Large Language Models (LLMs) have revolutionized natural language processing by learning from vast amounts of data to generate coherent and contextually relevant text~\cite{dubey2024llama, wang2024healthq, zhao2024towards}. Despite their impressive capabilities, a critical question persists: to what extent do these models memorize their supervised fine-tuning (SFT) data versus generalize to new, unseen inputs? While LLMs can generate plausible text, they also risk reproducing verbatim passages from their training datasets, leading to significant privacy and security concerns~\cite{carlini2021extracting}. Memorization occurs when a model outputs exact or near-exact replicas of its training data, which can result in the unintended exposure of sensitive information or violations of intellectual property rights.

Recent research indicates that memorization in LLMs during SFT does not affect all training data equally; instead, certain data points are significantly more prone to memorization, leading to a highly \emph{skewed} pattern. For instance, Xie et al.~\cite{xie2024memorization} found that LLMs often rely on memorization when solving logical reasoning tasks like Knights and Knaves puzzles. The models achieved high accuracy on training puzzles but struggled with slightly perturbed versions, suggesting that they memorized specific instances rather than learning underlying reasoning principles. This skewness means that while most data points are generalized over, a small subset contributes disproportionately to the overall memorization risk. Consequently, average-case analyses or mean memorization rates~\cite{carlini2021extracting, carlini2022quantifying, feldman2020does} fail to capture these worst-case scenarios where sensitive information may be leaked, akin to assessing system security based solely on average performance without considering rare but critical failures.

Furthermore, methods relying on sampling to estimate maximum memorization can be biased and may overlook rare but significant instances, particularly under practical constraints like limited sample sizes or computational resources~\cite{carlini2022quantifying, schwarzschild2024rethinking}. Prior studies have also artificially increased memorization by replicating training data multiple times (e.g., thousands of repetitions)~\cite{carlini2021extracting}, which is not reflective of real-world training settings. Such approaches may not accurately identify where memorization is most pronounced within a model or elucidate the factors that contribute to it. This limitation hampers effective comparison of memorization across different models or training configurations and underscores the need for methods that can detect and analyze memorization without unrealistic experimental setups.

In this work, we analyze skewed memorization in LLMs trained with SFT without relying on unrealistic experimental setups like replicating training data multiple times. Employing non-parametric statistical tests that handle skewed data distributions, we capture nuances overlooked by average-case analyses, providing more accurate assessments of worst-case scenarios and valuable insights into model behavior. By decomposing term-wise probabilities in the LLM generation process, our theoretical analysis reveals that data characteristics—such as similarity gaps and local data density—influence memorization likelihood; data points with very few or very many similar neighbors are less likely to be extensively memorized due to underfitting or high local entropy, respectively. Our experimental results show that memorization increases with more training epochs, even as overall loss decreases, indicating that prolonged training exacerbates memorization risks. Altering dataset composition or size significantly affects memorization patterns; mixing datasets or changing sizes leads to noticeable differences in which data points are memorized, highlighting the crucial role of dataset characteristics. By comparing our skewness-aware memorization metrics with traditional metrics like ROUGE~\cite{lin2004rouge} and Levenshtein distance~\cite{yujian2007normalized}, we connect our findings with established evaluation methods, making comparison of results across fields and eras handy.

% \section{Related Work}
\section{Methods}

We propose a \textit{prefix continuation} framework to quantify memorization in LLMs, measuring how many tokens a model recalls beyond a given prefix. To capture its skewed distribution, we use non-parametric sampling to estimate worst-case risks~\cite{feldman2020does, carlini2021extracting}. We analyze how memorization intensifies with training, varies with dataset size and composition, and correlates with embedding-space diversity, linking trends to scaling laws~\cite{kaplan2020scaling, hoffmann2022training}. Finally, we compare our metric to standard text similarity measures, demonstrating its effectiveness in capturing extreme memorization cases. Theoretical proofs and additional comparisons are provided in the Appendix.

\subsection{Estimating Memorization in LLMs}

Manual inspection of word completions often fails to capture memorization, as worst-case instances are systematically underestimated in small samples~\cite{carlini2021extracting, carlini2022quantifying}. Memorization in LLMs is highly skewed, requiring a distributional approach rather than reliance on mean-based measures. Denote cumulative distribution function $F = \mathbb{P}(N_{\text{pre}} \leq n)$, where $N_{\text{pre}}$ denotes random variable of consecutively recalled tokens from a training example beyond a given prefix, $n_{\text{pre}}[i]$ denotes the actual value at $i$-th data point. The probability that a sample of size $z$ fails to contain any of the top $k$ most memorized cases is given by: $\mathbb{P}(\max(n_1, \dots, n_z) < N_{(-k)}) = \frac{\binom{M - k}{z}}{\binom{M}{z}}$, which approximates how often extreme memorization cases are omitted under random sampling.

Since direct computation is impractical, we estimate $F$ via sampling-based methods. When $z \ll M$, uniform sampling with replacement closely approximates sampling without replacement, and the sample maximum follows: $\mathbb{P}(\max(n_1, \dots, n_z) < n) \approx \prod_{i=1}^{z} \mathbb{P}(N_{\text{pre}} < n)$, implying that smaller sample sizes systematically underestimate extreme memorization. To refine the probability of missing out worst cases, we use non-parametric resampling~\cite{efron1994introduction, feldman2020does}, which provides a non-parametric approximation of memorization distributions and has been widely applied in deep model analysis. To ensure robustness, we examine distributional properties such as normality, skewness, and kurtosis of the true population.

\subsection{Checkpoint-Level Memorization Dynamics Across Training}

Understanding how memorization evolves across training is essential for assessing its risks and developing mitigation strategies. We investigate memorization trends by analyzing how distributions change under three key conditions: increasing training epochs, reducing dataset size, and altering dataset composition. Our goal is to determine how these factors influence memorization and whether memorization intensifies in specific scenarios.

As training progresses, memorization is expected to increase, particularly in the upper quantiles, as overfitting leads to stronger memorization of training data~\cite{feldman2020does, carlini2022quantifying, tirumala2022memorization}. While overall loss decreases, extreme memorization cases often become more pronounced~\cite{zhang2021understanding}. Similarly, smaller training datasets provide fewer constraints for generalization, leading to greater reliance on memorization. This follows from established scaling laws in deep learning, where dataset size significantly impacts optimization dynamics and generalization ability~\cite{kaplan2020scaling, hoffmann2022training}. Although mean memorization may remain stable, its effects manifest in the upper tail of the distribution.

Beyond the size of the data set, composition also plays a crucial role in the behavior of memorization. If two datasets differ semantically, their combination modifies the embedding space, potentially altering memorization patterns. We explore whether such changes influence memorization by examining shifts in memorization distributions when subsets of datasets are merged.

To systematically measure these effects, we adopt a \textit{ distribution analysis} approach rather than relying solely on mean-based memorization metrics. The top-$k$ memorization curve provides a more sensitive indicator of extreme cases. To formally compare memorization distributions across training conditions, we employ \textit{non-parametric statistical tests}, ensuring robustness to distributional shifts without assuming specific parametric forms. Since training loss alone does not directly quantify memorization, empirical evaluation remains essential to characterize memorization trends and assess their impact across different training configurations.

\subsection{Analyzing Skewed Memorization: Identifying Highly Memorized Data Points}

We introduce a \textit{prefix continuation} approach to systematically measure memorization in LLMs. Given a training sequence $s$, we partition it into a prefix $r_{\text{pre}}$ of length $c$ and a suffix $r_{\text{ref}}$. The model, prompted with $r_{\text{pre}}$, generates a continuation $s$. The \textit{prefix match length} $n_{\text{pre}}$ is defined as: $n_{\text{pre}} = \max \{ k \mid {s}_{1:k} = r_{\text{ref},1:k} \}$, where ${s}_{1:k}$ is the first $k$ tokens of the generated output, and $r_{\text{ref},1:k}$ is the corresponding ground truth segment. The random variable $N_{\text{pre}}$ captures the distribution of memorization lengths across the dataset.

We seek to determine which training samples exhibit longer memorization. Let $i$ index training samples and $j$ denote token positions within the suffix. This allows us to analyze memorization both across different training samples and within individual sequences. A deeper examination of this structure provides insights into how autoregressive models generate tokens based on prior context, allows bounding of memorization length distributions, and facilitates comparisons between memorization metrics.

\begin{table}[t]
\small
\centering
\caption{Notation used in the memorization analysis.}
\begin{tabular}{p{1.2cm}p{6.3cm}}
\toprule
\textbf{Symbol} & \textbf{Description} \\
\midrule
$N$ & Maximum tested inference length \\
$M$ & Number of training samples \\
$R_b$ & Bayesian risk of incorrect recall \\
$D_{KL}$ & Kullback-Leibler divergence between learned and true distributions \\
$\pi_{M}(j|.)$ & Estimated token distribution at $j$ given prior tokens \\
$\pi(j|.)$ & True token distribution given prior context \\
$M_b$ & Bayes Optimal Classifier (BOC) for suffix prediction \\
$M_t$ & Term-wise BOC predicting token $j$ given prior tokens \\
$M_C$ & LLM at training checkpoint $C$ \\
$MI(J_{\text{pre}}, j)$ & Mutual information between memorized prefix and token $j$ \\
$Comb(j)$ & Set of all possible combinations of prior tokens before $j$ \\
\bottomrule
\end{tabular}
\end{table}

\subsubsection{Decomposing Memorization Patterns}

Since LLMs generate tokens sequentially, later outputs do not causally affect earlier ones~\cite{brown2020language}. This enables factorization of the probability of a consecutive match ending at $n_{\text{pre}}$:

% \begin{multline*}
% P(N_{\text{pre}} = n_{\text{pre}})\\
% = 
% \prod_{j=1}^{n_{\text{pre}}} P(j \text{-th term coincides with output} \\ 
% \mid \text{prev. correct tokens up to j-1}) \notag \\
% \quad \cdot P(n_{\text{pre}}+1 \text{-th term fails to coincide with output} \\ 
% \mid \text{prev. correct tokens up to } n_{\text{pre}}). 
% \end{multline*}

\begin{equation*}
\mathbb{P}(N_{\text{pre}}=n_{\text{pre}})= (\prod p_j^o) (1-p_{n[pre]}^o)
\label{decomposition}
\end{equation*}

where $p_k^o$ for each k is the probability that the model memorizes $k$-th block given $r_{pre}$ and given all previous memorization correct.
% this formulation is ugly but the previous is not quite precise in the meaning

Autoregressive models generate token $j+1$ based on all prior tokens, making $\mathbb{P}(j|r_{\text{pre}}, 1 \dots j-1)$ largely independent of whether these prior tokens were model outputs or ground truth inputs: $\prod_{j=1}^{n_{\text{pre}}} \mathbb{P}(j|r_{\text{pre}}, 1 \dots j-1) \cdot \mathbb{P}(n_{\text{pre}}+1^c | r_{\text{pre}}, 1 \dots n_{\text{pre}})$. .

If an approximately uniform token-wise memorization probability $p$ takes place, then the distribution follows a geometric pattern: $N_{\text{pre}} \sim \text{Geom}(1 - p)$. This assumption is reasonable in scenarios where token-wise loss is evenly distributed. However, a more general model incorporates position-dependent memorization probabilities. We denote correct memorization at token $j$ as $C(j|\text{input})$, where $C(j) = 0$ if correct and $1$ otherwise. We test whether memorization at $j$ depends on earlier memorization states using mutual information: $MI(C_{J_{\text{pre}}}, C_j) = H(j) - H(j | J_{\text{pre}})$. 

If $MI(J_{\text{pre}}, j) \approx 0$, memorization at token $j$ is independent of prior correct outputs, validating the geometric approximation. When this condition holds, we have $p_k=p_k^o$, can denote \ref{decomposition} as 
$(\prod^{n_{pre} } p_j) (1-p_{n}[pre])$. To further analyze memorization dynamics, we test cases where the per-token memorization probability follows a linear trend $p_j = \alpha j + p_0$. The expected prefix match length in this setting is:

$
\mathbb{P}(N_{\text{pre}} = n_{\text{pre}}) = p_0^{n_{\text{pre}}} \frac{\Gamma(n_{\text{pre}} +1+ \alpha/p_0)}{\Gamma(1 + \alpha/p_0)} (1 - p_{n_{\text{pre}}+1}).
$

This formulation allows modeling memorization behavior under different training conditions. Empirical verification is performed by plotting $\mathbb{P}(n_{\text{pre}} > j-1 | n_{\text{pre}} \geq j-1)$ as $p_j$, eliminating the need for term-wise inference.

\begin{remark}
Although some training samples exhibit more memorization than others, the probability of token $j$ being memorized is not necessarily dependent on the memorization of earlier tokens. This result holds under the assumption that memorization is driven primarily by local context and token-wise probability distributions.
\end{remark}

\subsubsection{Memorization and Embedding Diversity}

Memorization in LLMs varies across training samples, with certain sequences exhibiting longer recall. We hypothesize that memorization correlates with the diversity of possible continuations for a given prefix. If a prefix appears in multiple training examples but leads to varied completions, memorization is less likely due to high entropy in suffix prediction~\cite{feldman2020does, jagielski2022measuring, carlini2022quantifying}. Conversely, when a prefix consistently maps to a single completion, memorization is more probable~\cite{zhang2021understanding}. Prior work supports that LLMs are more likely to memorize deterministic mappings~\cite{tirumala2022memorization, brown2020language}, reinforcing the role of training data structure in memorization risk.

To formalize this intuition, we compare memorization under different modeling assumptions: (i) the \textit{Bayes Optimal Classifier (BOC)}, which minimizes classification risk given prior knowledge~\cite{shalev2014understanding}, (ii) the \textit{term-wise Bayes Optimal Classifier} ($M_t$), which optimally predicts each token given its context, and (iii) \textit{LLM inference dynamics}, which approximate these classifiers in practice~\cite{zhang2021understanding}. This comparison provides insight into when and why memorization occurs.

The \textit{Bayes Optimal Classifier} provides an upper bound on memorization by selecting the most probable suffix $s[1:n]$ given a prefix $r_{\text{ref}}$:

\begin{theorem}
    The maximum fraction of memorization no shorter than $n$ is achieved by the Bayes Optimal Classifier:
    \begin{equation*}
        s[1:n] = M_b(r_{\text{ref}}) = \arg\max_{S^n} \pi(S^n = r^n \mid r_{\text{ref}}),
    \end{equation*}
    with expected Bayesian risk:
    \begin{equation*}
        R = \sum_{r_{\text{ref}}} \left( 1 - \max_{S^n} \pi(S^n = s^n \mid r_{\text{ref}}) \right) \mathbb{P}(r_{\text{ref}}).
    \end{equation*}
\end{theorem}

This result establishes that memorization likelihood depends on the entropy of suffix distributions conditioned on the prefix. When suffix diversity is high, memorization is less likely, aligning with prior findings that memorization is more common in low-entropy regions of training data. To refine this intuition, we analyze memorization at the token level using a term-wise classifier.

\begin{definition}
    The \textit{term-wise Bayes Optimal Classifier} predicts the most probable next token given prior tokens:
    \begin{equation*}
        M_t(r_{\text{ref}}, \dots, j) = \arg\max_{s[j]} \pi(s[j] = r[j] \mid r_{\text{ref}}, \dots, j-1).
    \end{equation*}
\end{definition}

This formulation decomposes memorization into per-token probabilities. If memorization follows a token-wise process, then maximizing $P(s_j = r_j \mid r_{\text{ref}}, \dots, j)$ should yield the most likely continuation. In  unless there are at least a pair of data replicated at some length

\begin{corollary}
    The term-wise classifier satisfies:
    \begin{equation*}
        \mathbb{P}(s_j = r_j \mid r_{\text{ref}}, \dots, j) \leq \max p_j.
    \end{equation*}
\end{corollary}

\begin{theorem} We define the index of possible outcome from the generation process by g. If this generation sampling g is independent of data sampling z within all the indices sharing the same input $r[z]_j=r_{pre}[j]=r_{ref}\dots j$ $$s_j[z][g] \perp r_j[z] |r_{ref}\dots j$$ One can think of this as generation process sampling cannot see the data index. When the argmax of the term of greedy generation coincides with the true argmax, using a non-greedy generation sampling does not increase the expected rate of term-wise memorization. \begin{equation} \begin{split} P(s_j[g]==r_j|r_{ref}\dots j)\\ \leq p(r_j[z]=argmax_{r[j]} p(r_j|r_{ref}\dots j) ) \\ = \max p_j \end{split} \end{equation} \end{theorem}

If LLMs behave as $M_t$, their memorization should align with per-token likelihood optimization rather than full-sequence memorization. Under small-temperature inference, if $M_t$ converges to the empirical token distribution, then: $D_{\text{KL}}(\pi \parallel \pi_M) \rightarrow 0 \Rightarrow M_t \approx \pi$. Thus, optimizing LLM loss aligns with $M_t$ rather than $M_b$, a property observed in large-scale training~\cite{kaplan2020scaling}.

\begin{theorem}
    The LLM greedy decoding process approximates $M_t$ if and only if: $\arg\max_{\pi[M]} \pi_M(s[j]) = \arg\max_{\pi} \pi(r[j])$.
\end{theorem}

This condition is weaker than full distributional convergence ($\pi_M = \pi$), meaning that LLMs may behave similarly to term-wise classifiers even if they do not fully match empirical distributions. This partially explains why LLMs can generalize while still memorizing certain phrases.

Since term-wise memorization is data-dependent, we hypothesize that memorization length is inversely correlated with local embedding diversity. Given an embedding function $f$, we define the \textit{embedding similarity gap}: $\Delta S = S_{\text{full}} - S_{\text{input}}$, where $S_{\text{input}}$ measures mean cosine similarity of embeddings between prefixes, and $S_{\text{full}}$ measures similarity between full sequences. If $\Delta S$ is large, suffixes vary more than prefixes, leading to lower memorization likelihood.

\begin{remark}
    Training samples with highly similar prefixes but highly diverse full sequences exhibit lower memorization probability. This follows from the term-wise classifier formulation, as suffix variability increases classification entropy, reducing sequence-level memorization.
\end{remark}

To empirically validate this hypothesis, we measure memorization length $N_{\text{pre}}$ against $\Delta S$ across different training configurations. This enables an assessment of how embedding space structure influences memorization, a key factor in privacy-aware training strategies.

\section{Experiments and Results}

Our initial trails revealed an unexpected challenge: memorization proved remarkably difficult to detect in domain-specific datasets. This observation led us to examine how dataset characteristics, specifically blending and size, influence memorization patterns. Upon collecting a larger sample, we discovered that the distribution of memorization instances closely approximated a geometric distribution, guiding our research toward decomposition analysis.

\subsection{General Setup}

For the baseline case, we utilized Lavita-Medical-QA\footnote{\url{https://huggingface.co/datasets/lavita/medical-qa-datasets}}, a medical question-answering dataset from Hugging Face. Originally formatted as multiple-choice questions, we modified it into a single-answer question set, extracting 9,723 question-answer pairs to form a domain-specific dataset. To introduce diversity, we incorporated GPTeacher-General-Instruct\footnote{\url{https://huggingface.co/datasets/teknium/GPTeacher-General-Instruct}}, a GPT-4-generated modular dataset. We randomly sampled 9,723 records from GPTeacher to match the dataset size, ensuring a comparable structure. While Lavita-Medical-QA is domain-specific, GPTeacher provides open-domain question-answering data. Further details on both datasets are provided in the Appendix.

For modeling, we employed Llama-3.1-8b-Instruct\footnote{\url{https://huggingface.co/meta-llama/Llama-3.1-8B-Instruct}}, fine-tuning it in a two-stage process with LoRA before inference. Inference was conducted using vLLM~\cite{vllm}, a high-throughput, memory-efficient engine optimized for LLM inference, with a temperature of 0 to ensure deterministic outputs. Implementation details are elaborated in the Appendix.

To quantify memorization, we segmented each training sample into two parts: the first 100 characters were provided as input to the model, while the remaining portion was used for evaluation. The generated output was compared to the reference text on a word-by-word basis, treating words as lists separated by blank spaces. A key measure is the onsecutive word match $n_{\text{pre}}$ occurring immediately after the cut, accumulating from the first output block. 

\subsection{Checkpoint-Level Memorization Trends}

\begin{figure}
    \centering
    \includegraphics[width=1\linewidth]{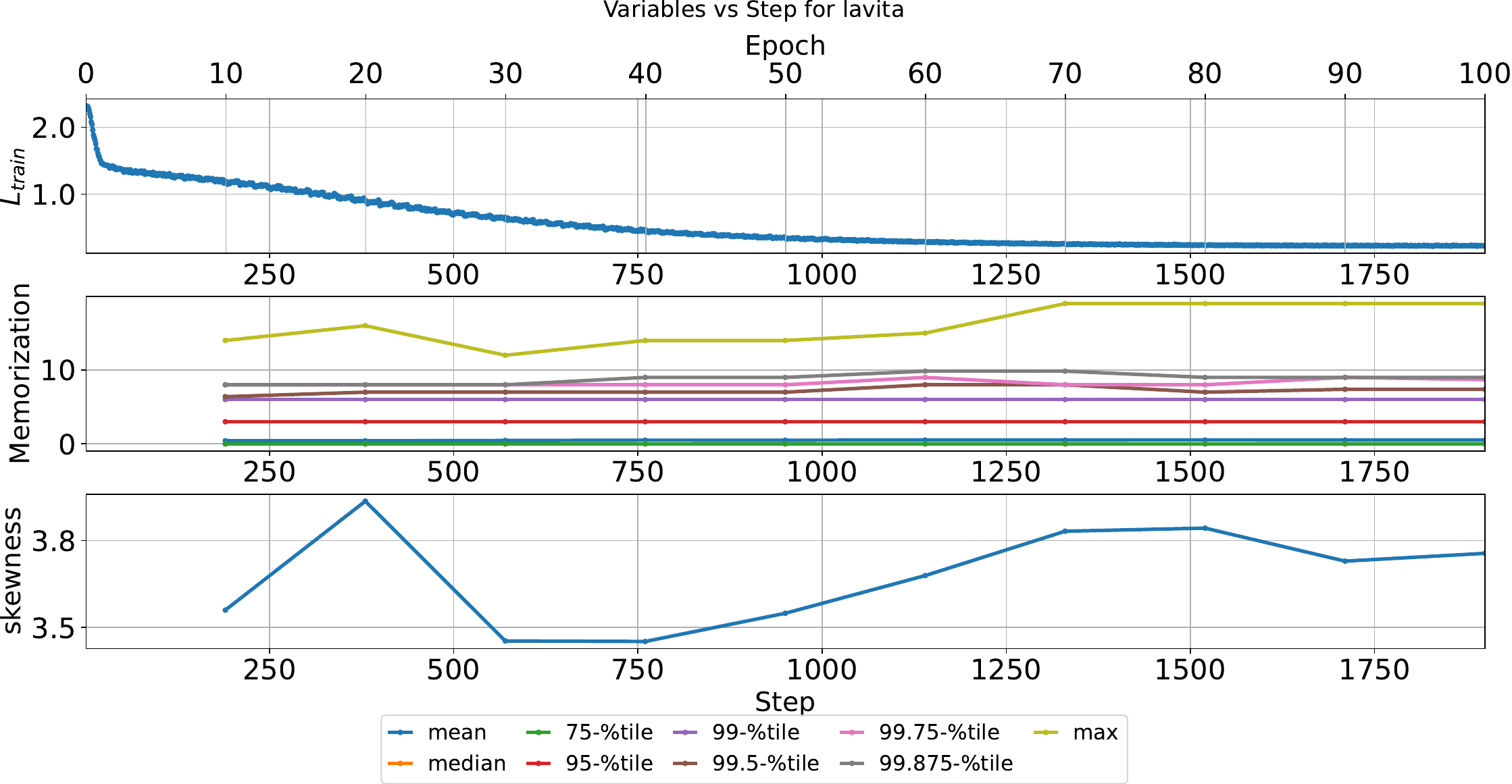}
    \caption{Memorization trends across training epochs in Lavita. Maximum and high-percentile memorization increase as loss decreases, but extreme cases appear early in training.}
    \label{fig:lavita_memorization}
\end{figure}

Memorization varies across training checkpoints and is influenced by dataset size, composition, and loss. Our analysis reveals that while memorization generally increases as loss decreases, worst-case memorization can emerge early in training. Figure~\ref{fig:lavita_memorization} illustrates this trend in the Lavita-Medical-QA dataset, showing increasing max memorization over 100 epochs.

% This refined visualization on the entire training dataset also shows that the maximum memorization rises without the necessity of having very low loss or high epochs. \ref{fig:lavita_memorization} shows max memorization with length greater than 10 even at epoch 10, when training loss remains above 1.0. Figure~\ref{fig:mixed_memorization} shows 
% maximum around 50 as at the first checkpoint at 10th epoch as well. This suggests that worst-case memorization may not be tightly bounded by loss, highlighting the limitations of using loss alone as a gauge of memorization.

\begin{figure}
    \centering
    \includegraphics[width=1\linewidth]{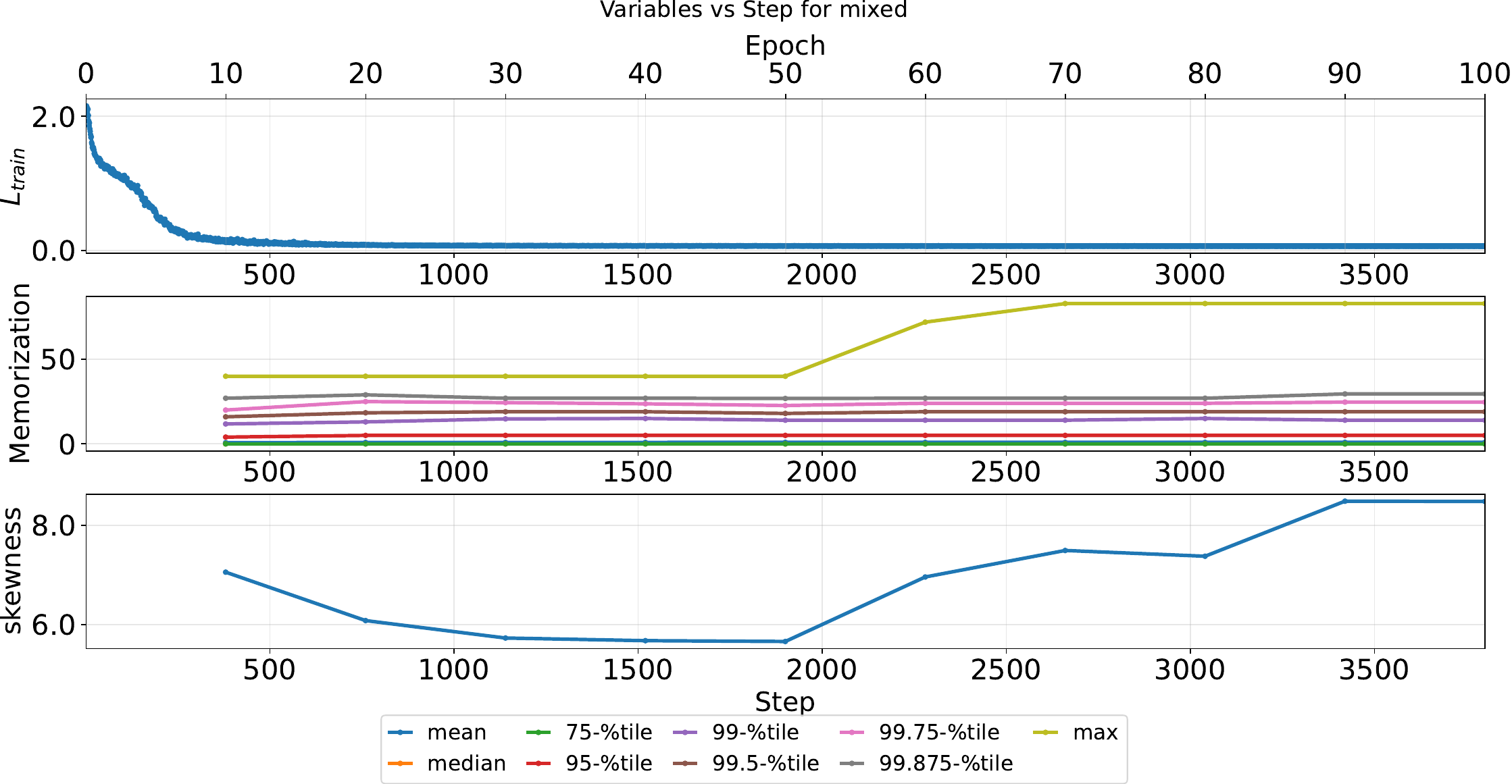}
    \caption{Memorization trends in the mixed dataset. While memorization patterns remain skewed, the overall distribution differs from Lavita, highlighting dataset-dependent memorization effects.}
    \label{fig:mixed_memorization}
\end{figure}

This comprehensive analysis of the full training dataset reveals that maximum memorization can increase independently of achieving extremely low loss or high epochs. As shown in Figure~\ref{fig:lavita_memorization}, memorization surpasses a length of 10 as early as epoch 10, despite the training loss remaining above 1.0. Similarly, Figure~\ref{fig:mixed_memorization} demonstrates a maximum memorization length of around 50 at the first checkpoint (epoch 10). These findings indicate that worst-case memorization is not strictly constrained by loss, underscoring the limitations of relying solely on loss as a proxy for memorization.

% To examine dataset effects on the memorization on the entire training-set, we introduce a second dataset by sampling 9,723 examples from the GPTeacher dataset, ensuring a comparable distribution in sequence lengths. We then construct a mixed dataset by replacing 200 randomly selected samples with corresponding examples from Lavita-Medical-QA, maintaining index consistency. The result shows much higher quantiles of memorization at same number of epochs compared to that of Lavita-Medical-QA--from the first checkpoint at epoch 10, the max memorization in the mixed set reached above 40 as compared to less than 16 in the first checkpoint of Lavita-Medical-QA. On the last checkpoint of 100 epochs, it reached above 50 for max memorization as compared to 16 in Lavita-Medical-QA.

To investigate the impact of dataset composition on memorization across the entire training set, we introduce a second dataset by sampling 9,723 examples from GPTeacher, ensuring a comparable distribution in sequence lengths. A mixed dataset is then constructed by replacing 200 randomly selected samples with corresponding examples from Lavita-Medical-QA while maintaining index consistency. Results indicate significantly higher memorization quantiles in the mixed dataset compared to Lavita-Medical-QA alone. At the first checkpoint (epoch 10), maximum memorization in the mixed dataset exceeds 40, whereas Lavita-Medical-QA remains below 16. By epoch 100, maximum memorization in the mixed dataset surpasses 50, while Lavita-Medical-QA remains at 16, demonstrating that dataset composition plays a crucial role in memorization dynamics.

\begin{figure}
    \centering
    \includegraphics[width=1\linewidth]{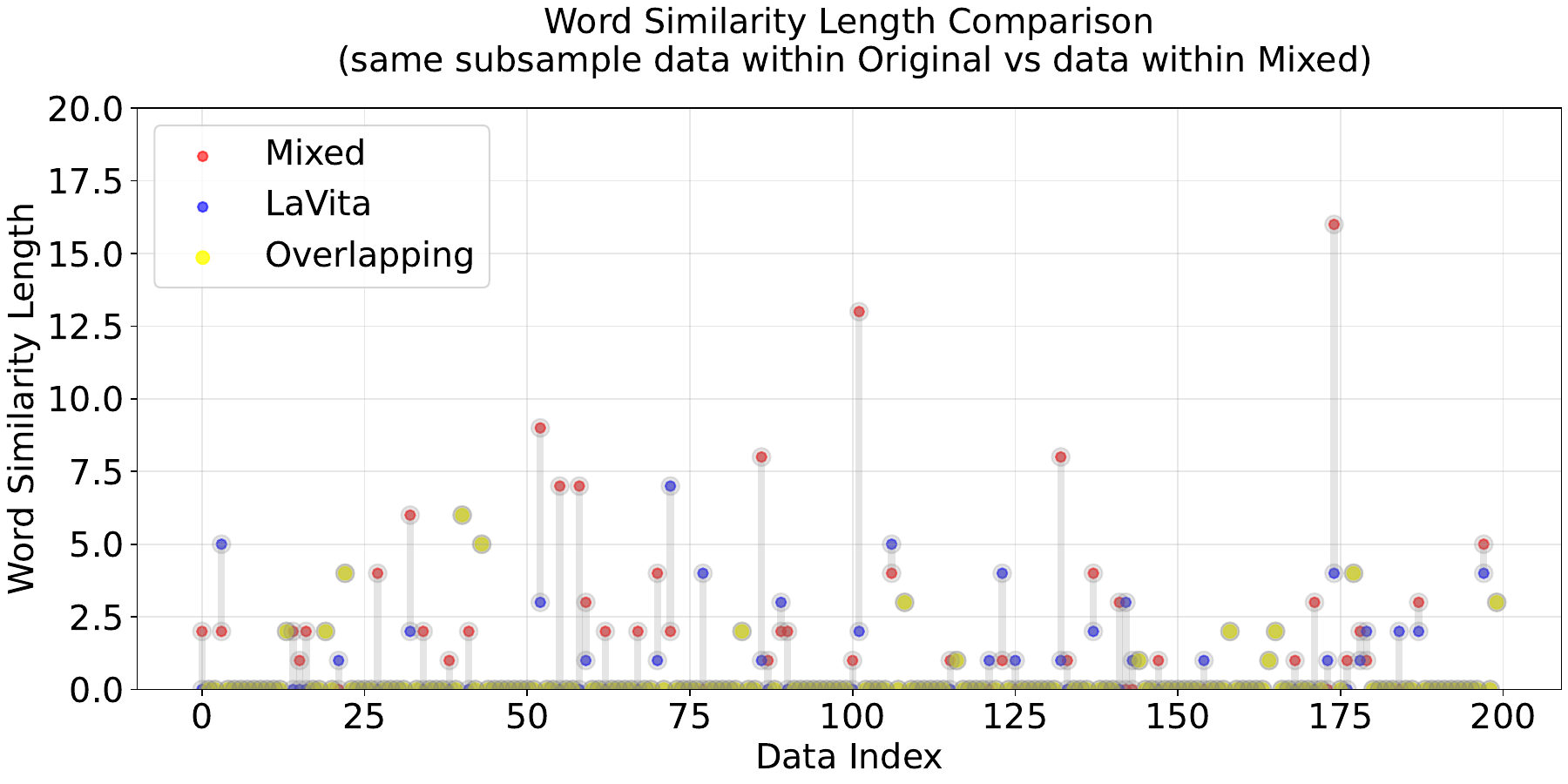}
    \caption{Memorization comparison for the same subset of samples trained in Lavita vs. the mixed dataset. Certain data points exhibit large memorization differences across contexts, emphasizing dataset-dependent effects.}
    \label{fig:comparison}
\end{figure}

We are also interested in the subsample we used that resides in both data. We compare the memorization of the 200 overlapping samples in the Lavita-Medical-QA and mixed datasets (Figure~\ref{fig:comparison}). A signed-rank test confirms a statistically significant difference in memorization between the two settings, reinforcing that memorization is not solely a property of an individual data point but also of the dataset context. Interestingly, while many samples exhibit similar memorization across datasets, a subset shows significantly different memorization lengths, with high variance and non-normal distribution. This highlights a potential risk: blending data from different domains can exacerbate worst-case memorization while altering per-instance memorization behavior in unpredictable ways.

% less important but intersting:
% The memorization distribution remains highly skewed at every training checkpoint, with a peak in skewness observed around epoch 20. As shown in Figure~\ref{fig:mixed_memorization}, the mixed dataset exhibits a different memorization distribution, underscoring the limitations of mean-based memorization assessments. Notably, the gap between the 99.875th percentile and maximum remains high [value], suggesting that extreme memorization cases persist even as overall trends shift.

\subsection{Scaling Laws in Memorization}

\begin{figure}
    \centering
    \includegraphics[width=1\linewidth]{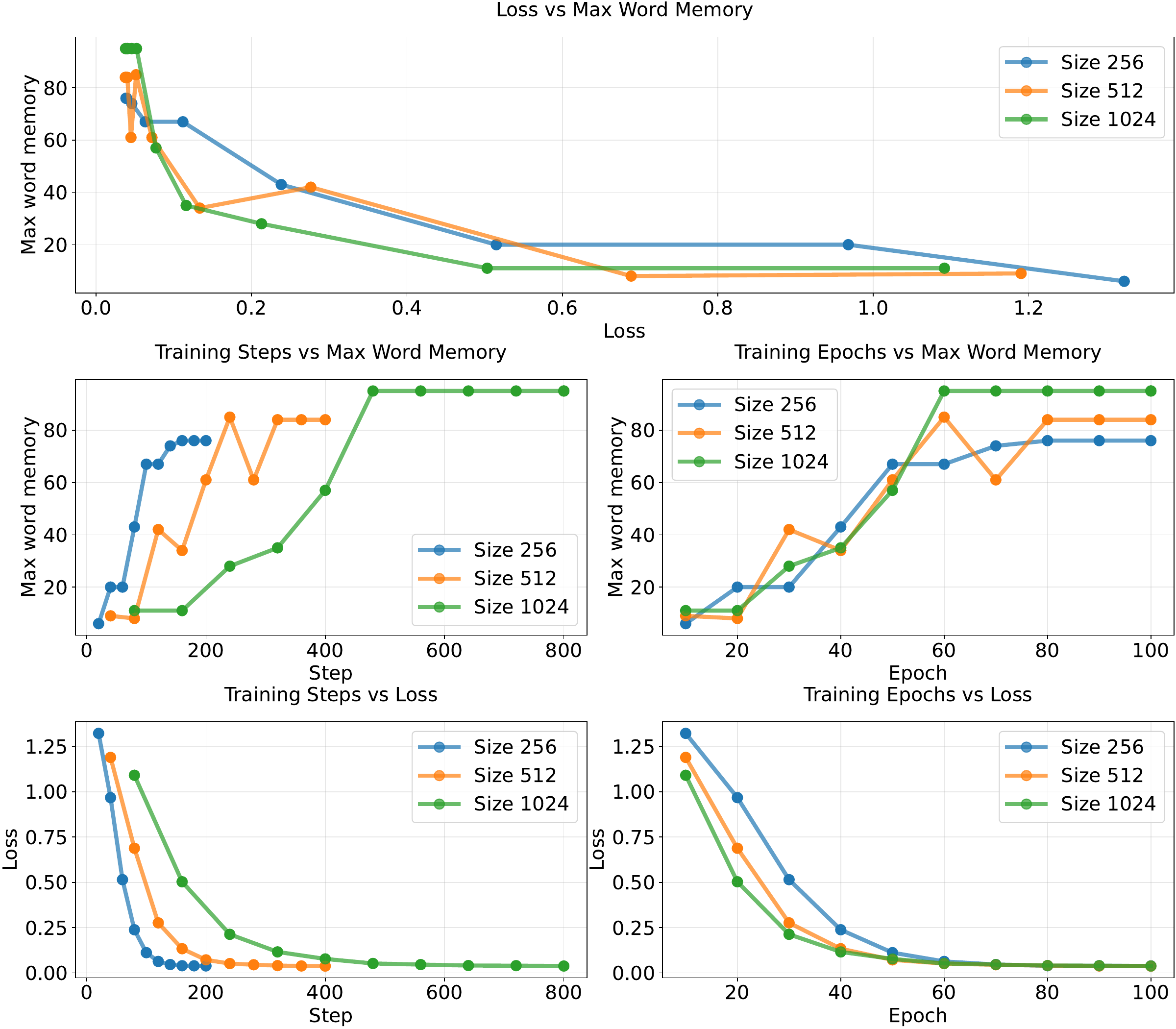}
    \caption{Memorization trends across smaller Lavita subsets. Despite fewer training steps per epoch, models rapidly reach high memorization levels, increasing risks for small-scale fine-tuning.}
    \label{fig:smaller_lavita}
\end{figure}

Expanding on prior research on scaling laws in deep learning~\cite{kaplan2020scaling, hoffmann2022training}, we investigate how dataset size influences memorization patterns. As described in the method section, we construct progressively smaller subsets of Lavita with sizes $2^8$, $2^9$, and $2^{10}$, training models using a consistent learning rate schedule. While smaller datasets yield lower per-step loss, systematic trends in memorization emerge when analyzed across epochs, loss progression, and memorization quantiles (Figure~\ref{fig:smaller_lavita}).

Our findings indicate that dataset size does not directly constrain maximum memorization but significantly affects how quickly memorization escalates throughout training. Smaller datasets impose fewer constraints on generalization, leading to a steeper rise in memorization as models overfit to the limited training samples. In the $2^8$ subset, maximum memorization reaches 70 within the first 100 training steps, with training loss dropping below 0.25. In contrast, in the $2^{10}$ subset, maximum memorization remains below 20 within the same interval, with loss staying above 1.0. 

This suggests that even when training on small datasets, models rapidly memorize individual samples despite achieving seemingly lower loss values. The effect is particularly pronounced in the smallest dataset, where limited training diversity allows models to memorize sequences much earlier in training. These results highlight the necessity of regularization techniques such as dropout, weight decay, or adversarial training when fine-tuning LLMs on domain-specific data with constrained sample availability. Moreover, they emphasize the need for more nuanced memorization assessments beyond loss-based evaluations, particularly in privacy-sensitive applications where early memorization can pose risks.

\subsection{Sampling Challenges in Skewed Memorization}

\begin{figure}
    \centering
    \includegraphics[width=1\linewidth]{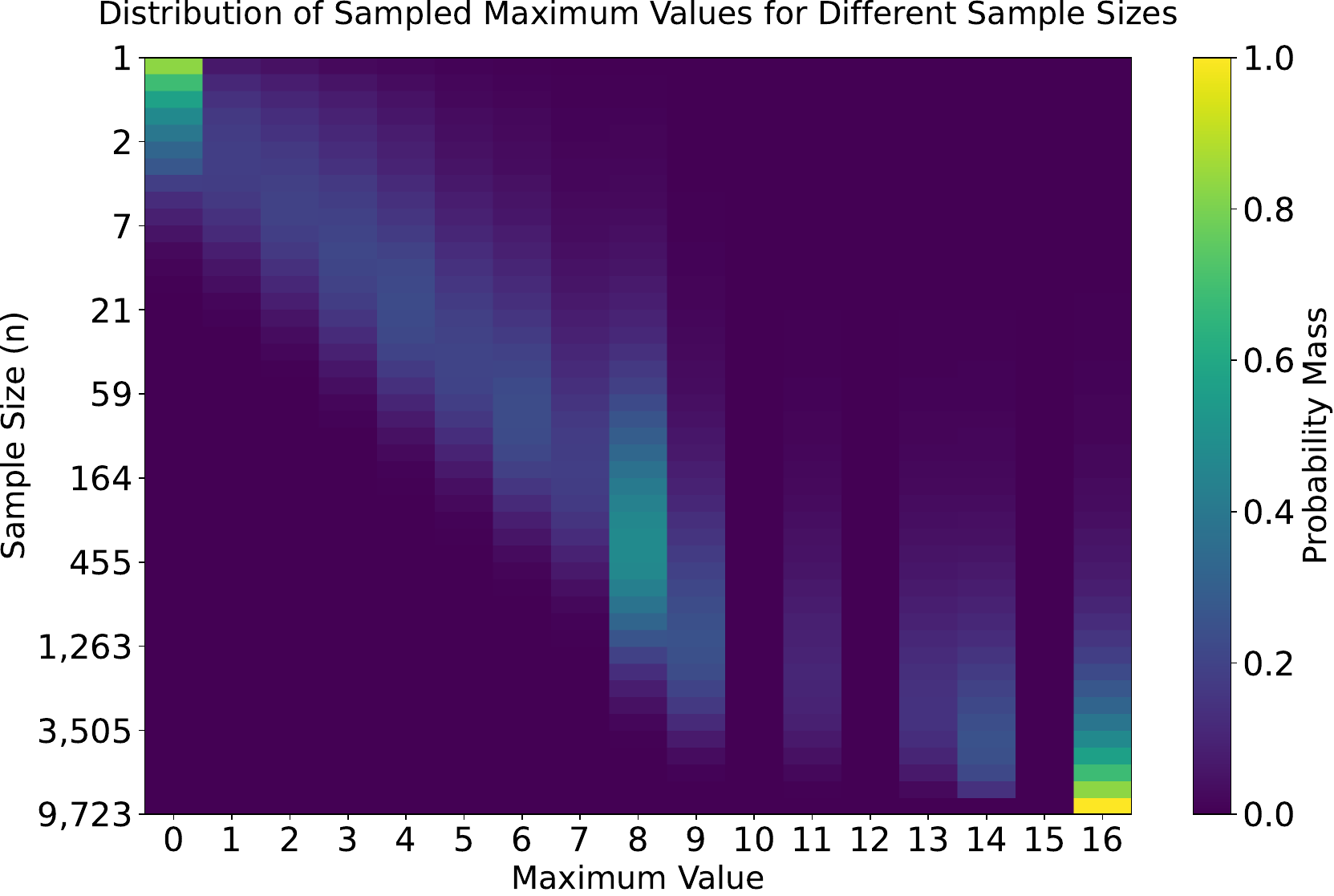}
    \caption{Probability of missing high-memorization instances as a function of sample size. Smaller samples fail to capture the distribution’s upper tail, leading to systematic underestimation of extreme cases.}
    \label{fig:prob_missing_max_density}
\end{figure}

Memorization in LLMs exhibits a highly skewed distribution, with extreme cases occurring infrequently but significantly. Our analysis reveals that while the average prefix match length remains low, the tail of the distribution extends substantially, indicating that certain samples experience much stronger memorization. Figure~\ref{fig:prob_missing_max_density} highlights this behavior. For example, if the sample size is $455$, there is a higher probability that the sampled max is $8$ as opposed to the true maximum $16$. showing that smaller sample sizes systematically underestimate worst-case memorization.

To quantify the reliability of detecting extreme cases, we evaluate the probability of missing high-memorization instances when sampling subsets of the training data. As expected, the likelihood of capturing maximum memorization scales with sample size, reinforcing the necessity of full-dataset scans for accurate estimation~\cite{sehanobishscalable}. This aligns with prior observations that worst-case memorization cannot be effectively characterized using small-scale evaluations~\cite{carlini2021extracting, carlini2022quantifying}.

These results emphasize the need for robust memorization assessment beyond mean-based measures. The observed variance in memorization across training checkpoints suggests that fine-grained distributional analysis is essential for understanding memorization dynamics and mitigating privacy risks in large-scale models.

\subsection{Decomposition}
 
\subsubsection{Memorization and Input Similarity}

As discussed in our method section, we hypothesize that memorization likelihood is influenced by data redundancy and embedding-space similarity. Specifically, if an input appears frequently in different contexts but leads to diverse continuations, memorization is less likely due to higher uncertainty in the suffix. Conversely, if a prefix consistently maps to a specific completion, memorization is more probable. To test this, we compute the mean embedding similarity of each input to all other inputs and compare it to the corresponding full-sentence similarity.

\begin{figure}
    \centering
    \includegraphics[width=1\linewidth]{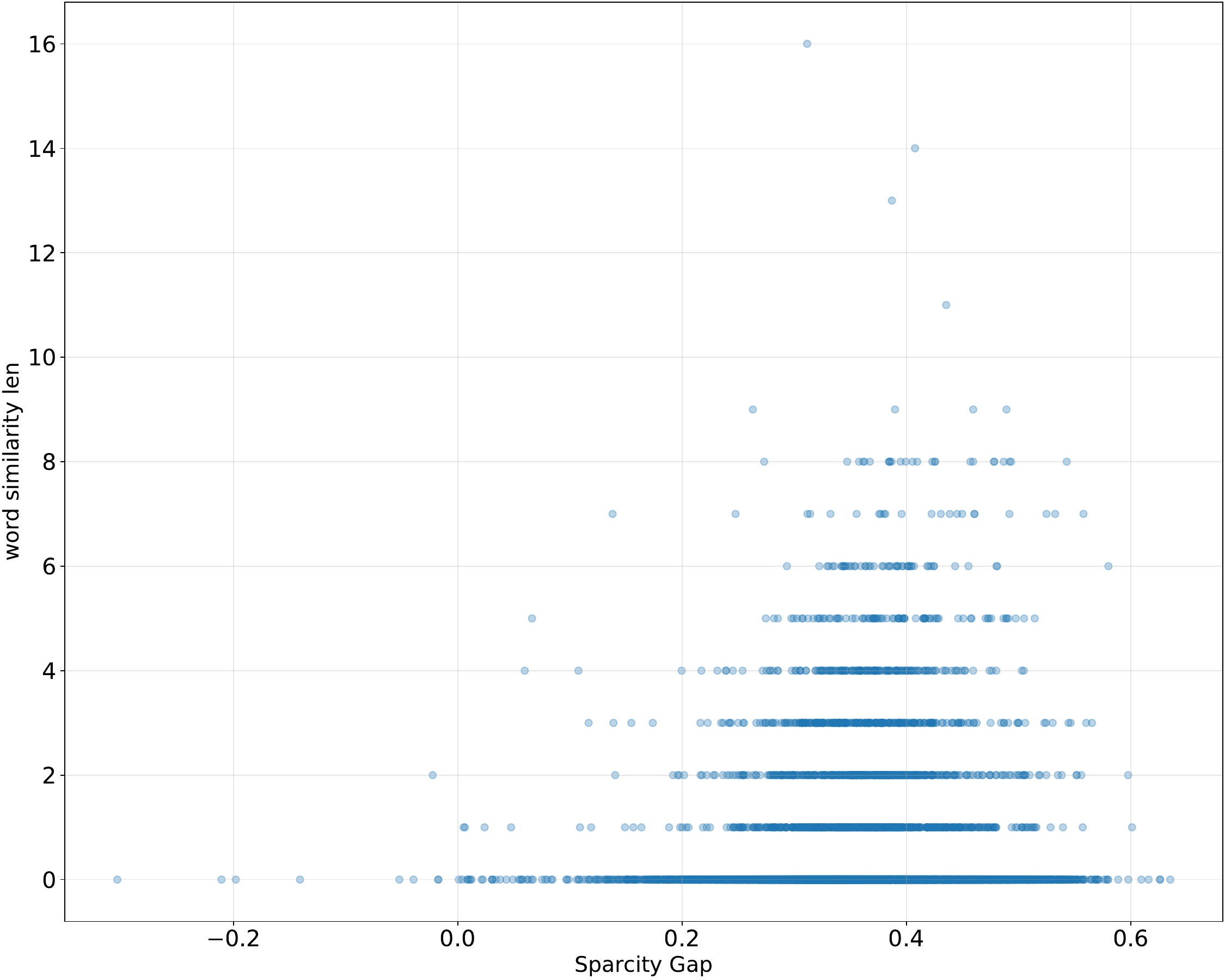}
    \caption{Memorization length ($n_{\text{pre}}$) against embedding similarity. Higher similarity to other training points correlates with increased memorization.}
    \label{fig:embedding_pattern}
\end{figure}

Figure~\ref{fig:embedding_pattern} shows $n_{\text{pre}}$ plotted against input and output similarities. The observed trend supports our hypothesis: higher input similarity correlates with stronger memorization, suggesting that data points residing in denser regions of the embedding space are more likely to be memorized. This trend is consistent across multiple training checkpoints in both the baseline and mixed datasets.

These findings reinforce that memorization is not solely a function of dataset size but also depends on how training samples are distributed within the learned representation. This aligns with previous studies on memorization risk in LLMs~\cite{carlini2021extracting, jagielski2022measuring} and suggests that dataset diversity at the embedding level could be a potential strategy for mitigating memorization.

\subsubsection{Decomposing Memorization at the Token Level}

To better understand how memorization propagates in LLMs, we analyze term-wise correctness probabilities by computing mutual information (MI) between memorization at different positions. Specifically, for each training sample $i$ and token position $j$, we compute a binary correctness matrix $c_{ij}$, where $c_{ij} = 1$ if the model correctly recalls the token at position $j$ given all previous tokens, and 0 otherwise. Using this matrix, we estimate the mutual information $MI(C_{J_{\text{pre}}}, C_j)$ for all valid prefixes $J_{\text{pre}} \subset j$.

\begin{figure}
    \centering
    \includegraphics[width=1\linewidth]{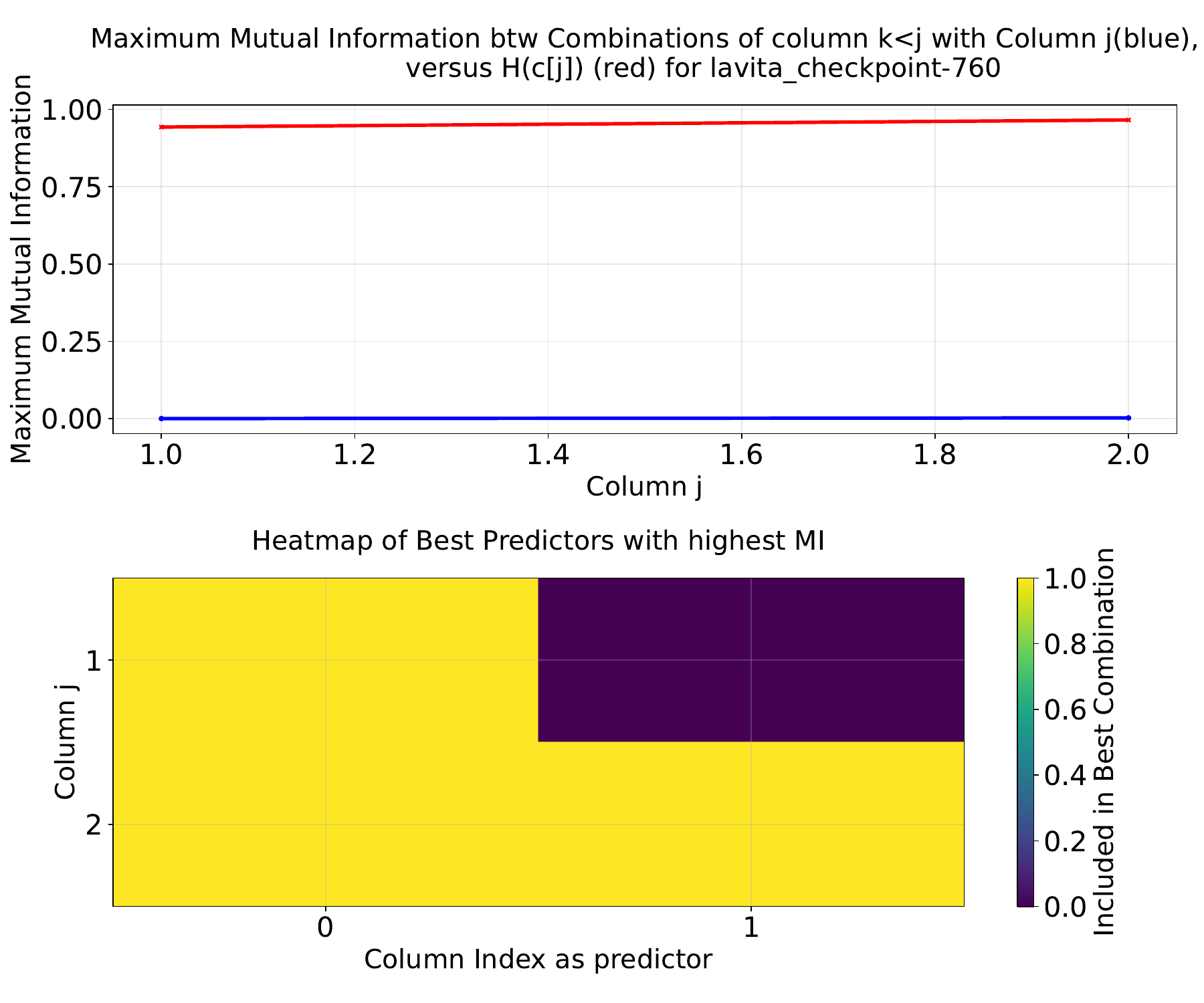}
    \caption{Mutual information between term-wise correctness and prefix correctness remains low across checkpoints, suggesting weak dependence.}
    \label{fig:mutual_info_is_low}
\end{figure}

\begin{figure}
    \centering
    \includegraphics[width=1\linewidth]{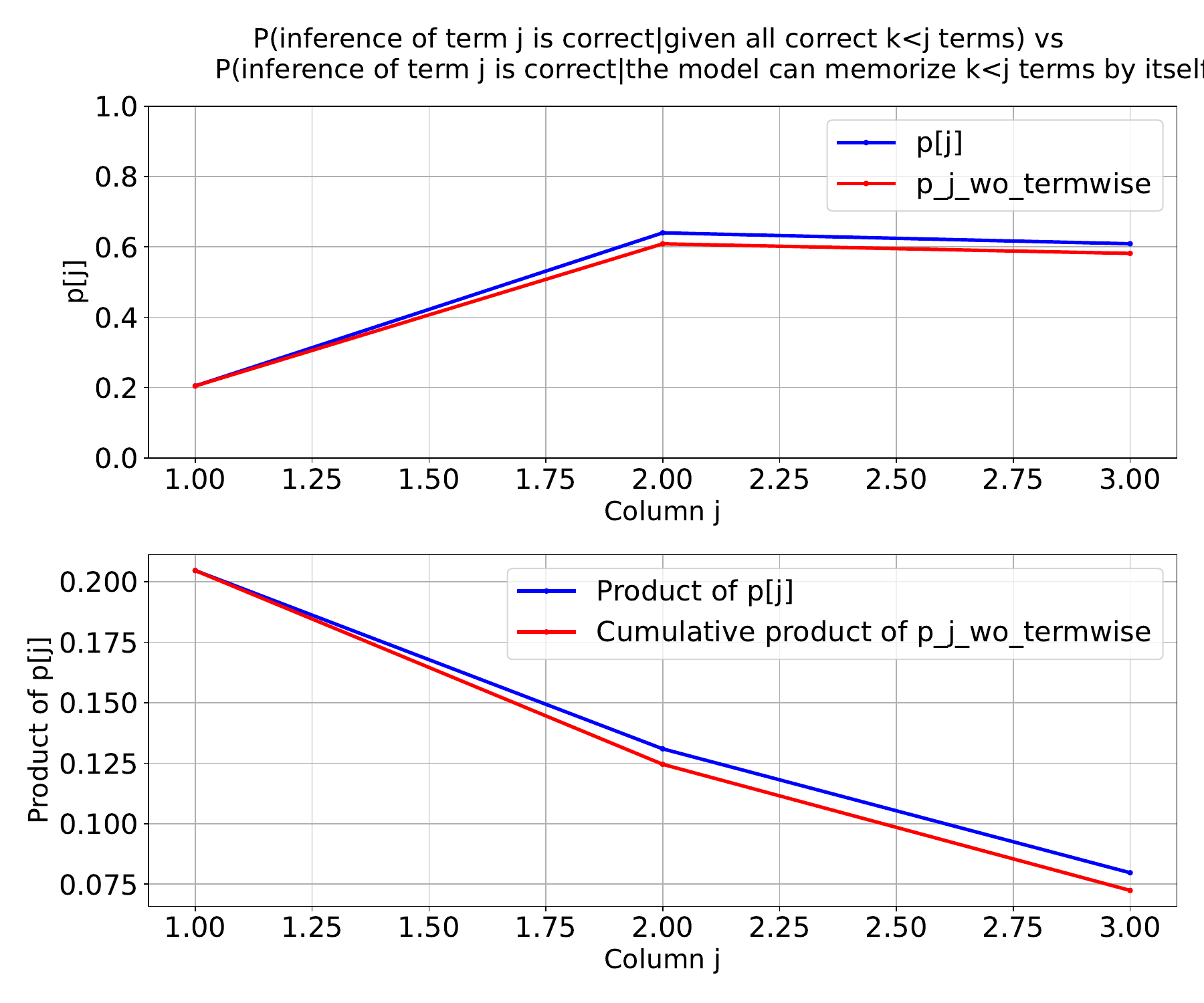}
    \caption{Conditional probability of correct token generation closely matches full probability estimates when MI is low.}
    \label{fig:cond_full_prob}
\end{figure}

Empirically, we find that $MI(C_{J_{\text{pre}}}, C_j)$ is consistently low across multiple training checkpoints in the baseline case (Figure~\ref{fig:mutual_info_is_low}). This suggests a weak dependency between correctness at $j$ and correctness at earlier positions, supporting the approximation: $P(N_{\text{pre}} \geq n) = 1 - F(n-1) \approx \prod_{j=1}^{n} p_j$. Since the cumulative product of per-token probabilities aligns closely with the observed memorization distribution (Figure~\ref{fig:cond_full_prob}), this validates our method section’s assumption that term-wise correctness probabilities approximate the memorization distribution. 

The independence of token prediction accuracy has critical implications for the reliability of model extraction attacks. The finding fundamentally undermines a key assumption in sequence extraction methodology: that successful recall of initial tokens can serve as a confidence indicator for subsequent extraction attempts. If correctness at position $j$ is independent of correctness at prior tokens, then testing memorization using a known prefix token is insufficient to infer correctness at subsequent tokens. This means that even if a model is tested to recall a specific word correctly, it does not imply high certainty for predicting the next token correctly. Consequently, assessing memorization should go beyond isolated token-level evaluations and consider full-sequence dependencies.

More broadly, despite the complexity of LLM architectures, our findings suggest that their token generation follows systematic probabilistic patterns, influenced by both training dynamics and structured loss guidance. This reinforces the idea that memorization can be characterized through well-defined statistical properties rather than requiring heuristic evaluations alone.

\section{Discussion}

Our findings highlight that prefix memorization in LLMs is rare but highly skewed, making worst-case instances difficult to detect without large samples. Dataset diversity influences memorization in two ways: it increases overall memorization but can reduce extreme cases within local input neighborhoods. Decomposing memorization into term-wise and sequence-level components further shows that token diversity within a sequence affects memorization likelihood, aligning with the structured nature of auto-regressive generation.

These insights have practical implications for detecting and mitigating memorization. Improved sampling strategies can enhance privacy risk assessment by capturing extreme cases more effectively. Additionally, increasing local diversity in training data may reduce unintended memorization while preserving model utility. Understanding how memorization relates to embedding-space structure and generation dynamics offers a path toward privacy-preserving model training and informed dataset curation.

\section{Future Work}

Several open directions merit further investigation. The connection between embedding similarity and memorization remains underexplored, particularly whether embedding collapse during training contributes to memorization intensity. Further research should examine whether memorization sensitivity extends beyond lexical similarity to factual relevance, potentially revealing how LLM embeddings encode structured knowledge. Understanding these mechanisms could inform future approaches for aligning LLMs with factual consistency while mitigating privacy risks.

\section{Conclusion}

This work presents a principled analysis of memorization in LLMs, revealing its skewed distribution, dependence on dataset diversity, and decomposition into term-wise and sequence-level components. By linking memorization to training dynamics and embedding-space structure, we provide both theoretical insights and practical interventions. These findings offer a foundation for refining memorization detection and improving privacy-preserving training strategies in large-scale language models.

\nocite{langley00}

\bibliography{ref}

\begin{thebibliography}{21}
\providecommand{\natexlab}[1]{#1}
\providecommand{\url}[1]{\texttt{#1}}
\expandafter\ifx\csname urlstyle\endcsname\relax
  \providecommand{\doi}[1]{doi: #1}\else
  \providecommand{\doi}{doi: \begingroup \urlstyle{rm}\Url}\fi

\bibitem[Brown et~al.(2020)Brown, Mann, Ryder, Subbiah, Kaplan, Dhariwal, Neelakantan, Shyam, Sastry, Askell, et~al.]{brown2020language}
Brown, T., Mann, B., Ryder, N., Subbiah, M., Kaplan, J.~D., Dhariwal, P., Neelakantan, A., Shyam, P., Sastry, G., Askell, A., et~al.
\newblock Language models are few-shot learners.
\newblock \emph{Advances in neural information processing systems}, 33:\penalty0 1877--1901, 2020.

\bibitem[Carlini et~al.(2021)Carlini, Tramer, Wallace, Jagielski, Herbert-Voss, Lee, Roberts, Brown, Song, Erlingsson, et~al.]{carlini2021extracting}
Carlini, N., Tramer, F., Wallace, E., Jagielski, M., Herbert-Voss, A., Lee, K., Roberts, A., Brown, T., Song, D., Erlingsson, U., et~al.
\newblock Extracting training data from large language models.
\newblock In \emph{30th USENIX Security Symposium (USENIX Security 21)}, pp.\  2633--2650, 2021.

\bibitem[Carlini et~al.(2022)Carlini, Ippolito, Jagielski, Lee, Tramer, and Zhang]{carlini2022quantifying}
Carlini, N., Ippolito, D., Jagielski, M., Lee, K., Tramer, F., and Zhang, C.
\newblock Quantifying memorization across neural language models.
\newblock \emph{arXiv preprint arXiv:2202.07646}, 2022.

\bibitem[Dubey et~al.(2024)Dubey, Jauhri, Pandey, Kadian, Al-Dahle, Letman, Mathur, Schelten, Yang, Fan, et~al.]{dubey2024llama}
Dubey, A., Jauhri, A., Pandey, A., Kadian, A., Al-Dahle, A., Letman, A., Mathur, A., Schelten, A., Yang, A., Fan, A., et~al.
\newblock The llama 3 herd of models.
\newblock \emph{arXiv preprint arXiv:2407.21783}, 2024.

\bibitem[Efron \& Tibshirani(1994)Efron and Tibshirani]{efron1994introduction}
Efron, B. and Tibshirani, R.~J.
\newblock \emph{An Introduction to the Bootstrap}.
\newblock Chapman \& Hall/CRC, 1994.

\bibitem[Feldman(2020)]{feldman2020does}
Feldman, V.
\newblock Does learning require memorization? a short tale about a long tail.
\newblock In \emph{Proceedings of the 52nd Annual ACM SIGACT Symposium on Theory of Computing}, pp.\  954--959, 2020.

\bibitem[Hoffmann et~al.(2022)Hoffmann, Borgeaud, Mensch, Buchatskaya, Cai, Rutherford, Casas, Hendricks, Welbl, Clark, et~al.]{hoffmann2022training}
Hoffmann, J., Borgeaud, S., Mensch, A., Buchatskaya, E., Cai, T., Rutherford, E., Casas, D. d.~L., Hendricks, L.~A., Welbl, J., Clark, A., et~al.
\newblock Training compute-optimal large language models.
\newblock \emph{arXiv preprint arXiv:2203.15556}, 2022.

\bibitem[Jagielski et~al.(2022)Jagielski, Thakkar, Tramer, Ippolito, Lee, Carlini, Wallace, Song, Thakurta, Papernot, et~al.]{jagielski2022measuring}
Jagielski, M., Thakkar, O., Tramer, F., Ippolito, D., Lee, K., Carlini, N., Wallace, E., Song, S., Thakurta, A., Papernot, N., et~al.
\newblock Measuring forgetting of memorized training examples.
\newblock \emph{arXiv preprint arXiv:2207.00099}, 2022.

\bibitem[Kaplan et~al.(2020)Kaplan, McCandlish, Henighan, Brown, Chess, Child, Gray, Radford, Wu, and Amodei]{kaplan2020scaling}
Kaplan, J., McCandlish, S., Henighan, T., Brown, T.~B., Chess, B., Child, R., Gray, S., Radford, A., Wu, J., and Amodei, D.
\newblock Scaling laws for neural language models.
\newblock \emph{arXiv preprint arXiv:2001.08361}, 2020.

\bibitem[Kwon et~al.(2023)Kwon, Li, Zhuang, Sheng, Zheng, Yu, Gonzalez, Zhang, and Stoica]{vllm}
Kwon, W., Li, Z., Zhuang, S., Sheng, Y., Zheng, L., Yu, C.~H., Gonzalez, J.~E., Zhang, H., and Stoica, I.
\newblock Efficient memory management for large language model serving with pagedattention.
\newblock In \emph{Proceedings of the ACM SIGOPS 29th Symposium on Operating Systems Principles}, 2023.

\bibitem[Lin(2004)]{lin2004rouge}
Lin, C.-Y.
\newblock Rouge: A package for automatic evaluation of summaries.
\newblock In \emph{Text summarization branches out}, pp.\  74--81, 2004.

\bibitem[Schwarzschild et~al.(2024)Schwarzschild, Feng, Maini, Lipton, and Kolter]{schwarzschild2024rethinking}
Schwarzschild, A., Feng, Z., Maini, P., Lipton, Z.~C., and Kolter, J.~Z.
\newblock Rethinking llm memorization through the lens of adversarial compression.
\newblock \emph{arXiv preprint arXiv:2404.15146}, 2024.

\bibitem[Sehanobish et~al.()Sehanobish, Choromanski, ZHAO, Dubey, and Likhosherstov]{sehanobishscalable}
Sehanobish, A., Choromanski, K.~M., ZHAO, Y., Dubey, K.~A., and Likhosherstov, V.
\newblock Scalable neural network kernels.
\newblock In \emph{The Twelfth International Conference on Learning Representations}.

\bibitem[Shalev-Shwartz \& Ben-David(2014)Shalev-Shwartz and Ben-David]{shalev2014understanding}
Shalev-Shwartz, S. and Ben-David, S.
\newblock \emph{Understanding machine learning: From theory to algorithms}.
\newblock Cambridge university press, 2014.

\bibitem[Tirumala et~al.(2022)Tirumala, Markosyan, Zettlemoyer, and Aghajanyan]{tirumala2022memorization}
Tirumala, K., Markosyan, A., Zettlemoyer, L., and Aghajanyan, A.
\newblock Memorization without overfitting: Analyzing the training dynamics of large language models.
\newblock \emph{Advances in Neural Information Processing Systems}, 35:\penalty0 38274--38290, 2022.

\bibitem[Wang et~al.(2024)Wang, Li, Huang, and Rahmani]{wang2024healthq}
Wang, Z., Li, H., Huang, D., and Rahmani, A.~M.
\newblock Healthq: Unveiling questioning capabilities of llm chains in healthcare conversations.
\newblock \emph{arXiv preprint arXiv:2409.19487}, 2024.

\bibitem[Xie et~al.(2024)Xie, Huang, Zhang, Yu, Chen, Lin, Li, Ghazi, and Kumar]{xie2024memorization}
Xie, C., Huang, Y., Zhang, C., Yu, D., Chen, X., Lin, B.~Y., Li, B., Ghazi, B., and Kumar, R.
\newblock On memorization of large language models in logical reasoning.
\newblock \emph{arXiv preprint arXiv:2410.23123}, 2024.

\bibitem[Yujian \& Bo(2007)Yujian and Bo]{yujian2007normalized}
Yujian, L. and Bo, L.
\newblock A normalized levenshtein distance metric.
\newblock \emph{IEEE transactions on pattern analysis and machine intelligence}, 29\penalty0 (6):\penalty0 1091--1095, 2007.

\bibitem[Zhang et~al.(2021)Zhang, Bengio, Hardt, Recht, and Vinyals]{zhang2021understanding}
Zhang, C., Bengio, S., Hardt, M., Recht, B., and Vinyals, O.
\newblock Understanding deep learning (still) requires rethinking generalization.
\newblock \emph{Communications of the ACM}, 64\penalty0 (3):\penalty0 107--115, 2021.

\bibitem[Zhao et~al.(2024)Zhao, Behari, Hughes, Zhang, Nagaraj, Tuyls, Taneja, and Tambe]{zhao2024towards}
Zhao, Y., Behari, N., Hughes, E., Zhang, E., Nagaraj, D., Tuyls, K., Taneja, A., and Tambe, M.
\newblock Towards a pretrained model for restless bandits via multi-arm generalization.
\newblock IJCAI, 2024.

\bibitem[Zheng et~al.(2024)Zheng, Zhang, Zhang, Ye, Luo, Feng, and Ma]{Zheng_LlamaFactory_Unified_Efficient_2024}
Zheng, Y., Zhang, R., Zhang, J., Ye, Y., Luo, Z., Feng, Z., and Ma, Y.
\newblock Llamafactory: Unified efficient fine-tuning of 100+ language models.
\newblock In \emph{Proceedings of the 62nd Annual Meeting of the Association for Computational Linguistics (Volume 3: System Demonstrations)}. Association for Computational Linguistics, 2024.
\newblock URL \url{https://arxiv.org/abs/2403.13372}.

\end{thebibliography}
\bibliographystyle{lib/icml2025}

%%%%%%%%%%%%%%%%%%%%%%%%%%%%%%%%%%%%%%%%%%%%%%%%%%%%%%%%%%%%%%%%%%%%%%%%%%%%%%%
%%%%%%%%%%%%%%%%%%%%%%%%%%%%%%%%%%%%%%%%%%%%%%%%%%%%%%%%%%%%%%%%%%%%%%%%%%%%%%%
% APPENDIX
%%%%%%%%%%%%%%%%%%%%%%%%%%%%%%%%%%%%%%%%%%%%%%%%%%%%%%%%%%%%%%%%%%%%%%%%%%%%%%%
%%%%%%%%%%%%%%%%%%%%%%%%%%%%%%%%%%%%%%%%%%%%%%%%%%%%%%%%%%%%%%%%%%%%%%%%%%%%%%%
\newpage
\appendix
\onecolumn

\section{Additional Experiment Setup Details and Generalization Explaination}

For the LoRA fine-tuning stage, we employed Llama-Factory \cite{Zheng_LlamaFactory_Unified_Efficient_2024}, an open-source and efficient framework designed for LLM Fine tuning. The following hyperparameters were consistently applied across all experiments during the LoRA fine-tuning process:

\begin{table}[h!]
\centering
\begin{tabular}{l c}
\toprule
\textbf{Parameter}                   & \textbf{Value}            \\ 
\midrule
\textbf{Base Model}                  & Llama3.1-8b-Instruct      \\ 
\textbf{Quantization Method}         & bitsandbytes              \\ 
\textbf{Quantization}                & int8                      \\ 
\textbf{LoRA Rank}                   & 256                       \\ 
\textbf{LoRA Alpha}                  & 512                       \\ 
\textbf{LoRA Dropout}                & 0.05                      \\ 
\textbf{Learning Rate}               & $1.00 \times 10^{-5}$     \\ 
\textbf{LR Scheduler}                & cosine                    \\ 
\textbf{Epochs}                      & 100                       \\ 
\textbf{Compute Type}                & bf16                      \\ 
\textbf{Attention Method}            & FlashAttention-2          \\ 
\textbf{Cutoff Length}               & 1024                      \\ 
\textbf{Batch Size}                  & 8                         \\ 
\textbf{Gradient Accumulation Steps} & 8                         \\ 
\textbf{Optimizer}                   & AdamW                     \\ 
\textbf{Warmup Ratio}                & 0.1                       \\ 
\bottomrule
\end{tabular}
\caption{Configuration Parameters for Llama3.1-8b-Instruct Training}
\label{tab:training_config}
\end{table}

We utilized vLLM \cite{vllm}, a high-throughput and memory-efficient engine designed for \emph{large language model (LLM) inference}. The LoRA fine-tuned Llama-3.1-8b-Instruct model was loaded in its full size, and the following generation configuration was applied within vLLM:

\begin{table}[h!]
\centering
\begin{tabular}{l c}
\toprule
\textbf{Parameter}                     & \textbf{Value}            \\ 
\midrule
\textbf{vLLM Version}                  & 0.63.0                   \\ 
\textbf{n (Number of output sequences)} & 1                        \\ 
\textbf{top\_k}                        & 1                        \\ 
\textbf{top\_p}                        & 1                        \\ 
\textbf{temperature}                   & 0                        \\ 
\textbf{max\_tokens}                   & 128                      \\ 
\textbf{repetition penalty}            & 1                        \\ 
\bottomrule
\end{tabular}
\caption{Configuration Parameters for vLLM}
\label{tab:vllm_config}
\end{table}

\section{Comparative Analysis with Existing Text Metrics}

To systematically compare memorization with standard text similarity metrics, we establish formal relationships between prefix memorization, common text comparison methods, and NLP evaluation criteria. 

\subsection{Notation and Definitions}

Table~\ref{tab:notation} summarizes the key notations used throughout our analysis.

\begin{table}[h]
\centering
\caption{Notation for Text Comparison Metrics}
\label{tab:notation}
\begin{tabular}{ll}
\toprule
\textbf{Symbol} & \textbf{Definition} \\
\midrule
$N_1$ & Length of generated output \\
$N_2$ & Length of reference text \\
$D_{\text{Levenshtein}}$ & Edit distance \\
$LCS$ & Longest Common Subsequence \\
ROUGE-L & ROUGE score based on $LCS$ \\
$n_{\text{pre}}|_N$ & Prefix match length, up to $N$ \\
ROUGE-$n$ & ROUGE score for $n$-grams \\
$n_{\max}$ & Max length of shared ordered $n$-grams \\
$c_i$ & 0-1 correctness at position $i$ \\
$d_w(s, r)$ & Weighted difference: $d_w(s, r) = \sum_1^n w_j c_j(s,r), w_j = w^{-j}$ \\
\bottomrule
\end{tabular}
\end{table}

\subsection{Handling Length Mismatch in Text Comparison}

A fundamental challenge in text similarity is handling cases where $N_1 \neq N_2$. We impose the constraint:
\[
n_{\text{pre}}|_{\max(N_1,N_2)} \leq \max(N_1, N_2).
\]
For missing indices, we set $c[i] = 1$ if one sequence has a token while the other does not, ensuring:
\[
d_w(s_1, s_3) \leq d_w(s_1, s_2) + d_w(s_2, s_3),
\]
which maintains the \emph{triangle inequality}, making $d_w$ a valid metric.

\subsection{Relations Between Memorization and NLP Metrics}

We establish the following hierarchical relationship:
\begin{multline*}
\text{Non-inplace matched word count} \geq LCS \geq n_{\max} \\
\geq \text{In-place matched $n$-gram length} \geq n_{\text{pre}} \geq \\
\min(N_1, N_2) - d_1 = \text{Total in-place matched words}.
\end{multline*}
Additionally:
\[
d_1 \geq \max(N_1, N_2) - LCS \geq \frac{D_{\text{Levenshtein}}}{2}.
\]

For weighted distances ($w > 2$), restricting to length $N$:
\[
d_r(s[i], r[i]) > d_r(s[i'], r[i']) \iff n_{\text{pre}}(s[i'], r[i']) > n_{\text{pre}}(s[i], r[i]).
\]

For $w=3$, we derive:
\begin{align*}
    d_3 &= \sum 3^{-j} c_j \leq \sum c_j \leq d_1, \\
    d_1 \cdot 3^{-n_{\text{pre}}} \frac{1}{1-1/3} &\geq \sum_{n_{\text{pre}}}^{\infty} 3^{-i} \geq d_3.
\end{align*}

\subsection{ROUGE Score Relations}

For ROUGE-L:
\[
\text{ROUGE-L} = \frac{LCS}{N_2} \geq \frac{n_{\max}}{N_2} \geq \frac{n_{\text{pre}}}{N_2}.
\]
For ROUGE-$n$:
\begin{align*}
    \text{Recall} &= \frac{\text{Count of summary $n$-grams in reference}}{\text{Total reference $n$-grams}}, \\
    \text{Precision} &= \frac{\text{Count of summary $n$-grams in reference}}{\text{Total summary $n$-grams}}.
\end{align*}

Denominators are $N_2 - n + 1$ and $N_1 - n + 1$, while the numerator satisfies:
\[
n_{\max} - n + 1 \geq n_{\text{pre}} - n + 1.
\]

\subsection{Implications for Memorization and Evaluation}

These results establish formal connections between memorization and standard NLP text similarity metrics, demonstrating that prefix memorization aligns with common evaluation measures while providing a more \emph{extreme-case-sensitive} assessment. This is crucial for detecting overfitting in models trained on small datasets or identifying privacy risks.

Since our memorization metrics satisfy fundamental \emph{metric properties}, they can be extended to generalization error analysis using clustering-based methods. Furthermore, standard evaluation metrics may systematically underestimate memorization risks, reinforcing the need for explicit memorization evaluation in LLMs.

All proposed metrics and relationships are implemented in our evaluation code, enabling direct application in diverse training setups.

\section{Proofs and Examples for Theorems and Properties}

\subsection{Probability of Memorization by One Token Ahead}

\begin{proof}
Since the sampling for generating a new token given the input is independent of sampling the actual data from the training set, we can express the probability of memorization at a given token as:

\begin{multline*}
    \mathbb{P}(\text{memorization at this token}|\text{input, all previous tokens}) \\
    =\sum_{y \in \mathcal{Y}}\mathbb{P}(\text{generated token}=y|\text{input, all previous tokens, actual token}=y) \\
    \mathbb{P}(\text{actual token}=y|\text{input, all previous tokens}) \\
    =\sum_{y \in \mathcal{Y}}\mathbb{P}(\text{generated token}=y|\text{input, all previous tokens}) \\
    \mathbb{P}(\text{actual token}=y|\text{input, all previous tokens}).
\end{multline*}

Since the sampling process for $\mathbb{P}(\text{actual token}=y|\text{input, all previous tokens})$ is determined by the training data distribution, we can enumerate $\mathcal{Y}$ and denote this probability as $p_i$. Similarly, the probability of generating token $y$ can be represented as $q_i$. The problem then simplifies to:

\[
\arg\max_{q} \sum p_i q_i
\]

subject to constraints:

\[
1 \geq p_i \geq 0, \quad 1 \geq q_i \geq 0, \quad \sum p_i = \sum q_i = 1.
\]

To prove that a greedy algorithm selects the token with the highest probability in the training data, we show that the optimal $q$ follows:

\[
q_i = \begin{cases}
1 & \text{if } i = \arg\max_i p_i \\
0 & \text{otherwise}.
\end{cases}
\]

Suppose a suboptimal $q \neq q^*$ exists. Then, there must be some index $i' = \bar{\arg\max}_i p_i$ with $q_{i'}>0$ and another index $i$ where $q_i^* < 1 - q_{i'}$. Constructing a new probability distribution by setting $q'_{\bar{\arg\max}[i] p[i]}=0$ and $q'_{{\arg\max}[i] p[i]} = q_{{\arg\max}[i] p[i]} +q_{\bar{\arg\max}[i] p[i]} $

Keeping other indices unchanged, we show that:

\[
\sum p_i q'_i > \sum p_i q_i.
\]

By iteratively adjusting $q$ in at most $|\mathcal{Y}|$ steps, we eventually reach $q^*$, proving the greedy selection process.
\end{proof}

Geometrically, this follows from differentiating $p_i q_i$ with respect to $q_i$, yielding $\nabla_q pq = p_i$. Since the function is linear in $q$, maximizing under an $L_1$ norm constraint is achieved by allocating all weight to the highest probability $p_i$.

\subsection{Bounding Memorization Using Weighted Correctness Scores}

\begin{proof}
Define two correctness vectors:

\[
C = (0,\dots,1,0,0,0\dots),
\]

where the first 1 appears at index $n$, and the rest are 0.

\[
c = (0, \dots, 0,1,1,1\dots),
\]

where the first 1 appears at index $n+1$ and is followed by consecutive ones. We show:

\[
\sum^{\infty} w^{-j}C_j > \sum^{\infty} w^{-j}c_j, \quad \text{for } w>2.
\]

By applying the geometric series sum formula, we compare cases where $n_{\text{pre}} = n-1$ and $n_{\text{pre}}' > n_{\text{pre}}$. For any correctness vector $c(n_{\text{pre}})$ corresponding to $n_{\text{pre}} = n-1$,

\[
\sum w^{-j}c_j(n_{\text{pre}}) > \sum^{\infty} w^{-j}C_j.
\]

Similarly, for $c(n_{\text{pre}}')$,

\[
\sum^{\infty} w^{-j}C_j > \sum w^{-j}c_j(n_{\text{pre}}').
\]

Thus, the order is preserved.
\end{proof}

\subsection{Inequalities with Edit Distance and ROUGE}

\begin{proof}
Using set inclusion, we establish the following sequence of inequalities:

\begin{align*}
\text{Non-inplace matched word count} &\geq \text{LCS} \geq n_{\max} \\
&\geq \text{Length of in-place matched n-gram} \geq n_{\text{pre}} \\
&\geq \max(N_1,N_2) - d_1 = \text{Number of in-place matched words}.
\end{align*}

For edit distance:

\[
d_1 \geq \max(N) - \text{LCS} \geq \frac{d_{\text{Levenshtein}}}{2}.
\]

Since an edit sequence from one string to another must at least remove non-LCS words and insert missing ones, we obtain:

\[
N_1 + N_2 - 2\text{LCS} \geq d_{\text{Levenshtein}},
\]

which bounds the edit distance.
\end{proof}

\subsection{Remarks on Conditional Probability and Memorization Strength}

\begin{remark}
Some data entries may be memorized more intensely than others, but the probability of memorization at a token, conditioned on previous tokens being correct, remains independent of specific token-wise memorization patterns.

Consider correctness sequences (where $i$ represents data and $j$ represents token position):

\[
\begin{array}{c}
01 \\
10 \\
11 \\
00
\end{array}
\]

Here, sequence 3 is more strongly memorized. The termwise correctness constraints are:

\[
\begin{array}{c}
0* \\
10 \\
11 \\
0*
\end{array}
\]

where * represents unrestricted positions. Computing $\mathbb{P}(2 \text{ correct} \mid 1 \text{ correct})$ shows:

\[
\mathbb{P}(2 \text{ correct} \mid 1 \text{ correct}) = \mathbb{P}(2 \text{ correct}).
\]

This provides a minimal counterexample, illustrating that termwise memorization probabilities do not contradict varying levels of memorization across data points.
\end{remark}

\begin{remark}
Choosing the highest probability term-by-term differs from selecting the highest probability contiguous subsequence.

Consider:

\[
\begin{array}{c}
000000 \\
000000 \\
123456 \\
132456 \\
142356 \\
152346 \\
162345
\end{array}
\]

A termwise greedy selection yields:

\[
1*****,
\]

memorizing at most 9 tokens but only fully memorizing one sequence. In contrast, selecting:

\[
0*****
\]

results in two fully memorized sequences. This discrepancy occurs when a less probable term is replicated over a longer sequence.

Thus, sequences with identical prefixes should not diverge significantly in subsequent terms to ensure consistent memorization behavior. This insight connects to the bounds of $R_b$ and $H$, motivating further experiments.
\end{remark}

\subsection{Expected Memorization in BOC vs. Termwise BOC}

\begin{proof}
If:

\[
\mathbb{E}[N_{\text{pre}}(\text{BOC})] > \mathbb{E}[N_{\text{pre}}(\text{T-BOC})],
\]

then they must differ at some index $j$. Given correct memorization up to $j-1$, the selection at $j$ in BOC leads to higher expected memorization than termwise BOC.

If the sequence extends at least $n_e$ beyond $j$, then:

\[
n_r \cdot n_j' > n_e - 1 + n_j[\max].
\]

Since $n_r, n_j'$ are integers, if $n_r \leq n_e$, then $n_j' \geq 2$, ensuring at least one pair is replicated in the training data at $|r_{\text{pre}}| + j + n_r$.
\end{proof}

\end{document}